\documentclass[conference]{IEEEtran}
\usepackage{times}

\usepackage[numbers]{natbib}
\usepackage{multicol}
\usepackage[bookmarks=true]{hyperref}

\usepackage{algorithm}
\usepackage{algpseudocode}
\usepackage{bbding}
\usepackage{multirow}
\usepackage{amsfonts}
\usepackage{amsmath}
\usepackage{graphicx}
\usepackage{amsmath}
\usepackage{float} 
\usepackage{graphicx}
\usepackage{booktabs}
\usepackage{tabularx} 
\usepackage{threeparttable}
\usepackage{makecell} 
\usepackage{cuted}      
\usepackage{caption}    

\begin{document}

\title{Act2Goal: From World Model To General Goal-conditioned Policy}

\author{Author Names Omitted for Anonymous Review. Paper-ID [add your ID here]}

\author{
Pengfei Zhou\textsuperscript{1,*},
Liliang Chen\textsuperscript{1,*},
Shengcong Chen\textsuperscript{1},
Di Chen\textsuperscript{1},
\\
Wenzhi Zhao\textsuperscript{1},
Rongjun Jin\textsuperscript{1},
Guanghui Ren\textsuperscript{1},
Jianlan Luo\textsuperscript{1,\dag}
\\
\textsuperscript{1}Agibot Research
\\
}

\maketitle

\renewcommand{\thefootnote}{\fnsymbol{footnote}} 
\footnotetext[1]{\textit{Equal contribution.}} \footnotetext[2]{\textit{Corresponding author.}}

\begin{abstract}
Specifying robotic manipulation tasks in a manner that is both expressive and precise remains a central challenge. While visual goals provide a compact and unambiguous task specification, existing goal-conditioned policies often struggle with long-horizon manipulation due to their reliance on single-step action prediction without explicit modeling of task progress.
We propose Act2Goal, a general goal-conditioned manipulation policy that integrates a goal-conditioned visual world model with multi-scale temporal control. Given a current observation and a target visual goal, the world model generates a plausible sequence of intermediate visual states that captures long-horizon structure. To translate this visual plan into robust execution, we introduce Multi-Scale Temporal Hashing (MSTH), which decomposes the imagined trajectory into dense proximal frames for fine-grained closed-loop control and sparse distal frames that anchor global task consistency. The policy couples these representations with motor control through end-to-end cross-attention, enabling coherent long-horizon behavior while remaining reactive to local disturbances.
Act2Goal achieves strong zero-shot generalization to novel objects, spatial layouts, and environments. We further enable reward-free online adaptation through hindsight goal relabeling with LoRA-based finetuning, allowing rapid autonomous improvement without external supervision. Real-robot experiments demonstrate that Act2Goal improves success rates from 30\% to 90\% on challenging out-of-distribution tasks within minutes of autonomous interaction, validating that goal-conditioned world models with multi-scale temporal control provide structured guidance necessary for robust long-horizon manipulation. Project page: \url{https://act2goal.github.io/}
\end{abstract}

\IEEEpeerreviewmaketitle



\section{Introduction}
\label{sec:introduction}

Learning robotic manipulation policies requires task specifications that are both expressive and precise. While natural language offers a flexible interface for diverse tasks, it often lacks the necessary granularity for fine-grained manipulation, a limitation that is more pronounced in complex, multi-stage scenarios. Visual goals offer a more precise alternative by directly encoding object configurations, spatial relations, and terminal constraints, avoiding linguistic ambiguity and explicit reward engineering.

Goal-conditioned policies (GCPs) map the current observation and a target visual goal directly to actions~\cite{kaelbling1993learning, liu2022goal, ding2019goal, gong2024goal}. While these methods perform well in short-horizon settings and exhibit reasonable generalization~\cite{black2023zero, tian2024predictive}, their performance degrades in long-horizon tasks. This limitation arises because standard GCPs operate via direct action prediction~\cite{reuss2023goal}, lacking an explicit representation of task progress, intermediate feasibility, or long-horizon consistency. This issue is further exacerbated when GCPs are trained on narrowly scoped demonstration data. Without an explicit model of visual transitions toward the goal, such policies tend to overfit demonstrated state--action mappings and must rely on dense supervision to compensate for the lack of structured intermediate guidance. This limitation becomes particularly pronounced in long-horizon or out-of-distribution settings, where maintaining coherent progress toward the goal requires reasoning beyond locally observed transitions.

\begin{figure}[t]
\centering
\includegraphics[width=\linewidth, trim=0 0 400 0, clip]{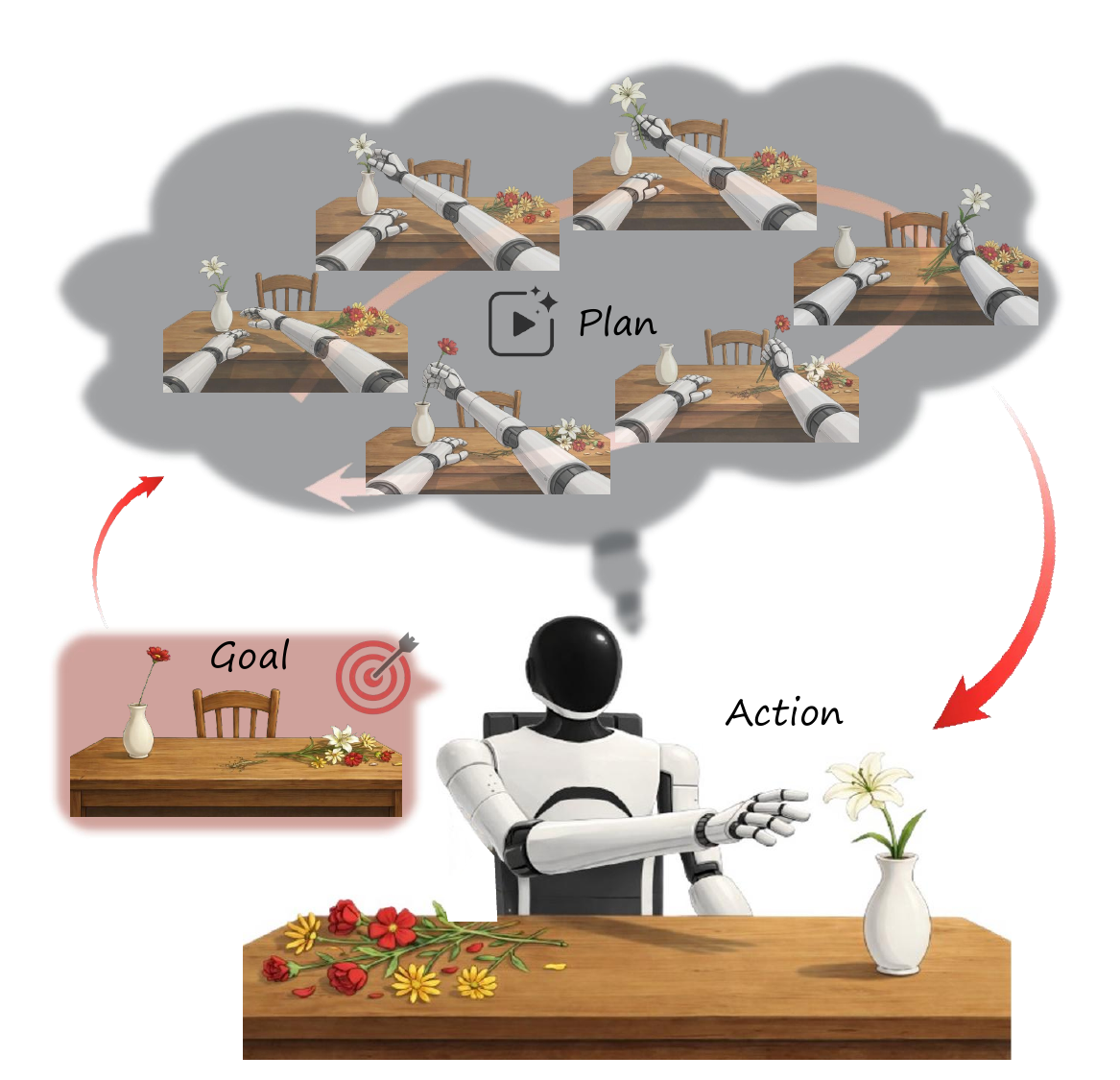}
\vspace{-0.4cm}
    \caption{\textbf{Method Overview.}
    The model receives a visual goal (left), imagines how to achieve it via a goal-conditioned world model (top), and executes the planned actions in the real world (right).}
    \label{fig:teaser}
    \vspace{-0.3cm}
\end{figure}

\begin{figure*}[t]
    \includegraphics[width=\linewidth, trim=0 25 0 0, clip]{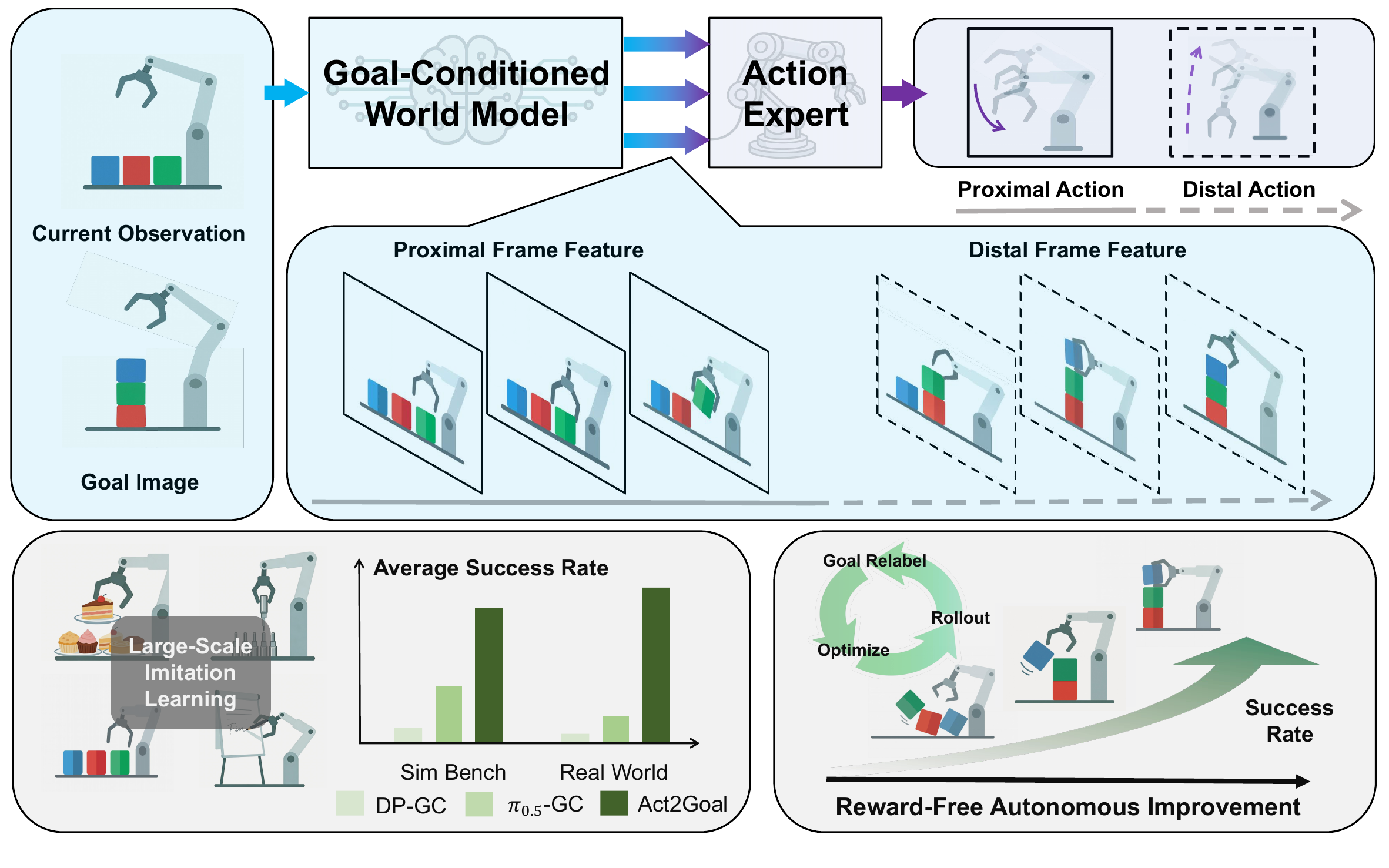}
    \captionof{figure}{\textbf{System Overview.} We propose Act2Goal, a goal-conditioned policy that integrates a visual world model with multi-scale temporal control to address long-horizon manipulation.
    After large-scale offline imitation learning, the model shows high performance on seen settings and strong generalization to unseen scenarios.
    The reward-free online autonomous improvement stage further improve model's performance through rollout-goal relabel-optimize loop.}
    \label{fig:system}
\end{figure*}


To generalize beyond demonstrations, GCPs must therefore explicitly model the visual dynamics required to reach a desired goal. Without such mechanism, the policy cannot distinguish between actions that make meaningful progress toward the goal and those that merely match local state–action correlations observed in demonstrations. 

Recent advances in world models provide a promising pathway to address these limitations~\cite{liu2024sora, hacohen2024ltx, agarwal2025cosmos}. Vision–language world models demonstrate that generative prediction of future visual states—conditioned on task instructions—can effectively support planning and decision-making \cite{hu2024video, wen2024vidman, liao2025genie}. 
Building on this insight, we consider a goal-conditioned analogue: goal-conditioned world model. Instead of predicting the future from high-level language, a goal-conditioned world model produces a plausible sequence of intermediate states bridging the gap between the current observation and a desired visual goal. 
This capability remedies a fundamental limitation of conventional GCPs by introducing an explicit representation of scene evolution over time, enabling a visually grounded and task-consistent trajectory toward the specified goal.

However, predicting a plausible visual path is only part of the solution. Executing long-horizon manipulation tasks reliably requires addressing a deeper control challenge. A general GCP must balance global consistency—maintaining fidelity to the long-term goal—with local reactivity—responding robustly to perturbations and correcting errors in closed-loop execution. Full-trajectory planning provides global coherence but is brittle under deviations; short-horizon control improves robustness but easily loses directional alignment in extended tasks. This inherent tension poses a fundamental barrier to deploying GCPs as general-purpose robotic controllers.

To bridge this gap, we introduce Act2Goal, a general goal-conditioned policy that integrates a goal-conditioned world model and a novel temporal decomposition mechanism termed Multi-Scale Temporal Hashing (MSTH). MSTH decomposes the generated visual trajectory into dense proximal frames for fine-grained control and sparse, horizon-adaptive distal frames that anchor the global plan. This multi-scale structure enables the policy to reason over long-horizon objectives while reacting quickly to local disturbances during closed-loop execution. 
We further couple the goal-conditioned world model with an action expert via layer-wise cross-attention, allowing intermediate visual representations to guide low-level motor control in an end-to-end differentiable architecture.

Initialized from large-scale human demonstration data, Act2Goal demonstrates strong zero-shot generalization across unseen objects, rearrangements, and environments. 
Beyond offline training, Act2Goal supports reward-free online autonomous improvement via Hindsight Experience Replay (HER)~\cite{andrychowicz2017hindsight}. 
By relabeling its own rollouts as additional goal-achieving trajectories and updating the policy efficiently through LoRA-based finetuning~\cite{hu2022lora}, the system rapidly adapts to new real-world scenarios without external supervision. In real-robot experiments, this enables Act2Goal to substantially improve performance on challenging long-horizon, out-of-distribution tasks, increasing success rates from 0.30 to 0.90 within minutes of autonomous interaction. These results validate the central premise of this work: that goal-conditioned world models, combined with multi-scale temporal reasoning, provide the structured intermediate guidance necessary for robust generalization and closed-loop execution in long-horizon manipulation.

Our main contributions are threefold. First, we present a novel end-to-end goal-conditioned policy that integrates a visual world model with motor control, enabling robust zero-shot generalization across unseen objects, environments, and goals. 
Second, we introduce Multi-Scale Temporal Hashing (MSTH), a temporal representation that decomposes trajectories into proximal and distal frames to balance long-horizon planning with closed-loop local control.
Third, we develop a reward-free online adaptation mechanism based on HER-style goal relabeling with LoRA-based finetuning, enabling rapid autonomous improvement on out-of-distribution tasks.


\section{Related Works}
\label{sec:relatedworks}
Our Act2Goal policy combines a goal-conditioned world model with an action expert capable of online autonomous improvement. Accordingly, we review related work in goal-conditioned policies, world models for robotic control, and online autonomous improvement.
 
\subsection{Goal-conditioned Policy}

Goal-conditioned policy learning aims to train agents to reach diverse formats of goals, e.g., visual goals~\cite{liu2022goal,ding2019goal,gong2024goal, jain2024go, kim2024goal,wang2023mimicplay,zhoutextit}, tracking points~\cite{wen2023any,bharadhwaj2024track2act} and motion field \cite{yin2025object}. Early works enhance imitation learning by relabeling goals~\cite{ding2019goal, andrychowicz2017hindsight}, while others extend reinforcement learning with structured goal representations~\cite{gong2024goal, davidson2025goals}. Recent advances further explore long-horizon reasoning~\cite{zhang2025chain}, keyframe-based planning~\cite{zhang2025chain}, and goal generation via program synthesis~\cite{davidson2025goals, ying2025assessing}.

Among them, GoalGAIL~\cite{ding2019goal} incorporates HER to learn from suboptimal or state-only demonstrations, improving sample efficiency in imitation settings. CoA~\cite{zhang2025chain} proposes generating action sequences in reverse from goal keyframes to maintain long-horizon consistency in manipulation.

These methods, however, typically rely on explicit goal supervision or struggle to align current observations with distant goals. Act2Goal addresses these limitations by introducing a goal-conditioned world model to simulate structured visual trajectories and proposing Multi-Scale Temporal Hashing (MSTH) to support consistent and efficient long-horizon planning, enabling better generalization in unseen tasks.


\subsection{World Model for Robotic Control}

World models have become a powerful tool in robotic control, enabling agents to simulate environmental dynamics~\cite{ha2018world, hafner2019dream, hafner2023mastering}, generate synthetic data for training~\cite{zhao2025high, liu2025robotransfer}, or serve as learned simulators to guide policy learning~\cite{jiang2025enerverse, guo2025ctrl}. Recent works further combine world models with action experts (AEs) to form policy planning systems~\cite{huang2025enerverse, liang2025video, wen2024vidman, hu2024video}, where the world model provides future state feature and the AE predicts actions accordingly.

Among these, GE-Act~\cite{liao2025genie} adopts a bi-model architecture with a world model predicting future visual feature based on language instruction and a transformer-based planner generating actions. WorldVLA~\cite{cen2025worldvla} jointly predicts vision and action in a unified latent space, aiming for tighter vision-action alignment.

Different from prior works, Act2Goal leverages a purely vision-based goal-conditioned world model to guide policy learning with structured visual trajectories. To the best of our knowledge, this is the first work to integrate a world model into goal-conditioned policy learning.

\subsection{Online Autonomous Improvement}

To enhance policy adaptability during deployment, recent studies explore online improvement via either interactive imitation learning such as DAgger~\cite{ross2011reduction, kelly2019hg, luo2025precise} or in-context learning (ICL)~\cite{shah2025mimicdroid, sridhar2025ricl}. However, DAgger-style methods require frequent expert intervention, while ICL methods do not update model weights and often show limited performance on complex tasks.

An alternative line of work leverages Hindsight Experience Replay (HER)~\cite{andrychowicz2017hindsight} and its extensions~\cite{yang2021mher, schramm2023usher, luo2023relay}, which relabel transitions by replacing original goals with achieved states, then optimize the policy using reinforcement or imitation learning. While these methods reduce the need for explicit rewards, they still rely on complex reward relabeling or external reward signals.

Act2Goal takes a further step toward fully autonomous online improvement. By combining HER-style relabeling with efficient LoRA-based finetuning, it enables direct policy adaptation from self-collected rollouts, without any task rewards or human annotations. This yields a lightweight, fully self-supervised update mechanism suitable for real-world deployment.

\section{From World Model To General Goal-conditioned Policy} 
\label{sec:method}

Our framework is designed to address two key challenges in long-horizon goal-conditioned manipulation: aligning the action policy with high-level goal semantics, and maintaining planning efficiency over extended time scales. As shown in Figure~\ref{fig:system}, we tackle the first challenge by introducing a \textbf{Goal-Conditioned World Model (GCWM)} to guide the policy with imagined visual futures, providing rich, temporally coherent representations. To address the second challenge, we propose \textbf{Multi-Scale Temporal Hashing (MSTH)}, which enables the policy to focus on both short-term execution and long-term goal awareness through structured temporal abstraction.

Our learning process consists of three stages. Stage~1 performs joint training of the GCWM and action expert to align their representations. Stage~2 focuses on action adaptation, further improving the policy’s performance. Stage~3 introduces optional autonomous improvement, allowing the model to self-adapt in novel scenarios during deployment. We describe each component in detail below.


\begin{figure}[t]
\centering
\includegraphics[width=\linewidth, trim=0 850 590 0, clip]{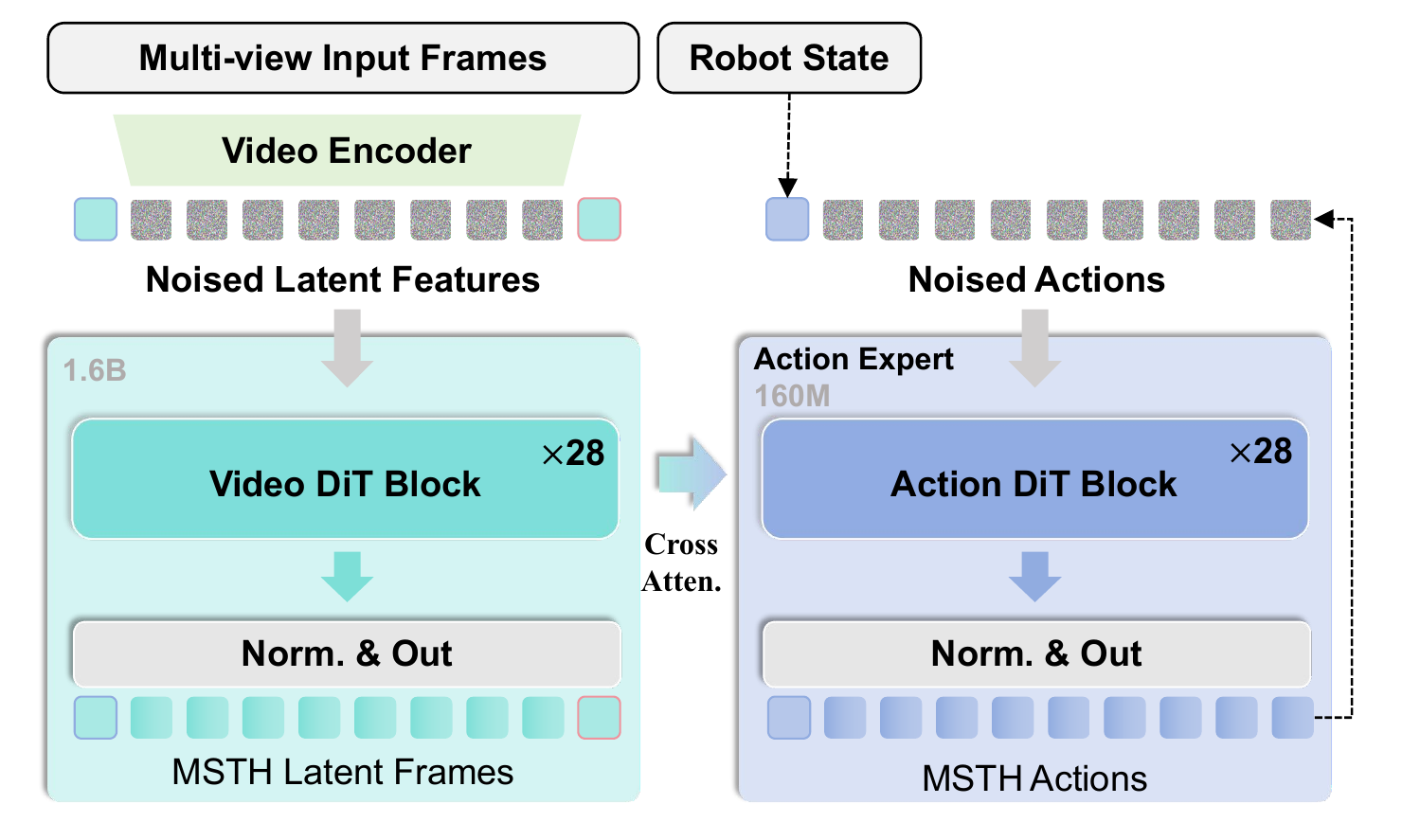}
\vspace{-0.7cm}
    \caption{\textbf{Model Architecture.}
    This figure presents the network architecture of Act2Goal model. On the left, multi-view input frames, including current observation and goal, are encoded into latents via a video encoder and concatenated with noisy latents, then refined into MSTH latent frames through Video DiT blocks. On the right, the robot state and multi-scale features from the world model are fed via cross-attention into isomorphic Action DiT blocks, generating MSTH-structured actions.}
    \label{fig:arch}
    \vspace{-0.3cm}
\end{figure}

\subsection{Goal-Conditioned World Model-Guided Policy }
As illustrated in Fig. \ref{fig:arch}, our goal-conditioned world model builds upon the Genie Envisioner architecture with key modifications tailored for goal-conditioned policy learning \cite{liao2025genie}. We introduce a goal visual condition that is concatenated with the current observation along the hidden states sequence, while removing all language-conditioning components to create a purely vision-based model.

Our goal-conditioned world model employs a continuous flow matching approach for generative modeling. The process can be abstracted as learning a transformation from random noise to structured visual sequences conditioned on both current observations and goal states:
\begin{equation}
\label{eq9}
z_{pred} = f_{\theta}(z_t, z_g, \epsilon),
\end{equation}
where $z_t$ and $z_g$ are VAE-compressed latents of the current observation and goal respectively, $\epsilon$ represents random noise inputs, which have the same shape as \(z_{pred}\), and $f_{\theta}$ is the flow matching model that generates the latent frames between current observation and goal.

During inference, the model progressively refines the noisy latents through a deterministic flow process:
\begin{equation}
\label{eq10}
z^{(n+1)} = z^{(n)} + \frac{1}{N}v_{\theta}(z^{(n)}, z_t, z_g),
\end{equation}
where $v_{\theta}$ is the learnt vector field that guides the denoising process over multiple steps $n = 0, 1, \dots, N-1$. The completed latent frames can be decoded into visual states using the VAE decoder.

Following the GE-Act implementation, our action expert employs a network architecture that is isomorphic to the world model, featuring the same number of DiT blocks but with reduced network width \cite{liao2025genie}. The action trajectory is predicted using a flow matching process conditioned on both the proprioceptive state $c_p$ and the multi-scale features $c_w$ from the world model.

The action prediction process can then be formulated as:
\begin{equation}
\label{eq12}
a_{pred} = g_{\phi}(c_w, c_p, \zeta),
\end{equation}
where $g_{\phi}$ is the action flow matching model, $\zeta$ are noise inputs for action generation, $c_w = \{h_{\text{world}}^1, \dots, h_{\text{world}}^L\}$ represents the layered transition features, and $c_p$ is the proprio state condition. The flow matching process during inference for actions follows an iterative refinement:
\begin{equation}
\label{eq13}
a^{(n+1)} = a^{(n)} + \frac{1}{N}u_{\phi}(a^{(n)}, c_w, c_p),
\end{equation}
where $u_{\phi}$ is the learned vector field for action generation.

\subsection{Multi-Scale Temporal Hashing for Visual State and Action}

In our framework, the \textit{Multi-Scale Temporal Hashing} (MSTH) mechanism jointly guides the goal-conditioned world model and the action expert, enabling a consistent yet flexible multi-scale temporal abstraction. Given a total imagined trajectory length $K$, a proximal horizon $P$, and a vision sampling stride $r$, MSTH partitions the future trajectory into two segments.

The proximal segment consists of high-frequency short-horizon visual states
$\{ s_{t+kr} \}_{k=1}^{P/r}$, which capture fine-grained local dynamics.
The distal segment contains $M$ sparsely sampled visual states
$\{ s_{t+d_m} \}_{m=1}^M$, where the indices $d_m$ are determined by logarithmic spacing:
\begin{equation}
d_m = P + \left\lfloor \frac{K - P}{\log(M+1)} \cdot \log(m+1) \right\rfloor, \quad m = 1, \dots, M .
\end{equation}
This logarithmic sampling results in increasing temporal intervals as the horizon extends, providing coarse but goal-aligned long-term guidance.

The predicted action sequence follows the same multi-scale structure, with an important distinction from vision. Proximal actions are predicted at \emph{every} timestep,
$\{ a_{t+1}, a_{t+2}, \dots, a_{t+P} \}$, enabling dense motor control even when visual states are subsampled by stride $r$. In contrast, distal actions
$\{ a_{t+d_m} \}_{m=1}^M$ are aligned with the distal visual states and serve as long-horizon guidance.
During deployment, only the proximal actions are executed, while distal predictions remain latent and guide long-term goal adherence.

\subsection{Two-Stage Offline Training}
To endow Act2Goal with strong generalization capabilities, we first train the model through large-scale offline imitation learning. The training process consists of two main stages, designed to ensure that the transition trajectory prediction objective aligns closely with the ultimate goal of guiding action planning.

In the first stage, we fine-tune a pre-trained world model to adapt it for the transition trajectory prediction task of generating multi-view video frames following the MSTH distribution between initial observation and goal condition. To enhance the alignment between transition trajectory prediction and action planning objectives, we jointly train both the transition trajectory prediction task and the action generation task using flow matching.

The training objective for the visual generation component follows the flow matching loss formulation:


\begin{equation}
\label{eq14}
\begin{split}
\mathcal{L}_v = \mathbb{E}_{t \sim U(0,1), z_0, z_1, z_t, z_g} \big[ \| v_{\theta}(t, \phi_t(z), z_t, z_g) \\
- (z_1 - z_0) \|^2 \big]
\end{split}
\end{equation}


\begin{equation}
\label{eq15}
\begin{split}
\mathcal{L}_a = \mathbb{E}_{t \sim U(0,1), a_0, a_1, c_w, c_p} \big[ 
\| u_{\phi}(t, \psi_t(a), c_w, c_p) \\
- (a_1 - a_0) \|^2 \big].
\end{split}
\end{equation}

The joint training objective combines both losses with a balancing coefficient $\lambda$. In our experiment, we set $\lambda = 0.1$:

\begin{equation}
\label{eq16}
\mathcal{L}_{\text{stage1}} = \mathcal{L}_v + \lambda \cdot \mathcal{L}_a.
\end{equation}

This joint optimization ensures that the goal-conditioned world model learns to generate visual trajectories that are not only visually plausible but also actionable, creating a strong foundation for the subsequent policy learning stage.

In the second training stage, we employ behavioral cloning to fine-tune the entire model end-to-end using only the action flow matching loss $\mathcal{L}_{\text{stage2}} =\mathcal{L}_a$, further strengthening its action planning capabilities. This stage focuses on aligning the complete pipeline—from visual perception to action execution—with expert demonstrations. The gradients from the action loss propagate through both the action generation components and the goal-conditioned world model, allowing the visual representations to be optimized specifically for action planning.

This two-stage offline training approach enables Act2Goal to acquire robust world understanding and action generation capabilities that transfer effectively to unseen environments and tasks.

\begin{algorithm}[t]
\caption{Act2Goal Online Autonomous Improvement}
\footnotesize
\begin{enumerate}
  \item Initialize replay buffer $\mathcal{B}$ and LoRA parameters $\phi$
  \item \textbf{While} performance not converged:
  \begin{enumerate}
    \item Execute policy for one episode, storing $(o, c_p, a, o')$ at each step
    \item \textbf{For each} transition $(o, c_p, a, o')$ in episode:
    \begin{enumerate}
      \item $\mathcal{B} \leftarrow \mathcal{B} \cup \{(o, c_p, a, o')\}$
    \end{enumerate}
    \item \textbf{If} $|\mathcal{B}| \geq N$:
    \begin{enumerate}
      \item \textbf{For} $k = 1$ to $K$:
      \begin{enumerate}
        \item Sample batch $\{(o,c_p,a,o')\} \sim \mathcal{B}$
        \item $g' \leftarrow o'$ \quad // Relabel achieved visual state as goal
        \item $\mathcal{L} = \mathbb{E}[\| \pi_{\theta}(o, c_p, g') - a \|^2]$ \quad 
        \item $\phi \leftarrow \phi - \alpha \nabla_{\phi} \mathcal{L}$
      \end{enumerate}
      \item $\mathcal{B} \leftarrow \emptyset$
    \end{enumerate}
  \end{enumerate}
\end{enumerate}
\end{algorithm}

\subsection{Online Autonomous Improvement}
While the model exhibits strong generalization capabilities after offline imitation learning, achieving high performance on novel tasks, environments, objects, and motion control chains remains challenging when deployed on physical robots—a common limitation of imitation learning-based policies. To address this, we introduce an online autonomous improvement framework based on Hindsight Experience Replay (HER), enabling autonomous performance enhancement during deployment \cite{andrychowicz2017hindsight}.

As illustrated in Algorithm 1, the system operates as follows: for a given expected goal condition, the model infers and executes actions. Each inference step—consisting of the starting observation latent, initial pripro state, output action, and resulting observation latent after execution—is automatically collected in a replay buffer on the edge device. Crucially, regardless of whether the transition successfully achieves the intended goal, we automatically relabel the goal condition as the robot's observation at the end of the inference step, eliminating the need for manual labeling of transitions. When the replay buffer reaches a predetermined capacity threshold, we perform a fixed number of end-to-end training iterations using the buffer transitions, following the same approach as Stage 2 of offline imitation learning. To ensure efficient on-device training, only the additionally introduced LoRA layers in the Act2Goal model are updated, while the base model parameters remain frozen \cite{hu2022lora}. After completing the fixed iteration training, the replay buffer is cleared, and the system continues collecting new data through rollouts, repeating this cycle until performance meets expectations. We set maximum inference counts for different tasks; exceeding this threshold triggers an automatic robot reset. During real-robot experiments, the only required human intervention is manually restoring or modifying the experimental setup when necessary.

\section{Experiments} 
\label{sec:experiments}
Our experiments aim to systematically evaluate the effectiveness of the proposed \textsc{Act2Goal} model across both offline and online settings. We assess the model's ability to generalize to novel scenarios, and to autonomously improve itself through interactions during deployment. Specifically, we study three key aspects: (1) the generalization capability of policies trained offline using imitation learning, measured under both in-domain (ID) and out-of-domain (OOD) scenarios; (2) the model’s capacity for online autonomous improvement, enabled by lightweight on-device fine-tuning of LoRA layers with self-relabelled transitions; and (3) the contribution of our proposed \textit{Multi-Scale Temporal Hashing} (MSTH) mechanism, evaluated through qualitative analysis and ablations on both ID and OOD tasks.

\begin{figure*}[t]
    \centering
    \includegraphics[width=\linewidth, trim=0 360 0 0, clip]{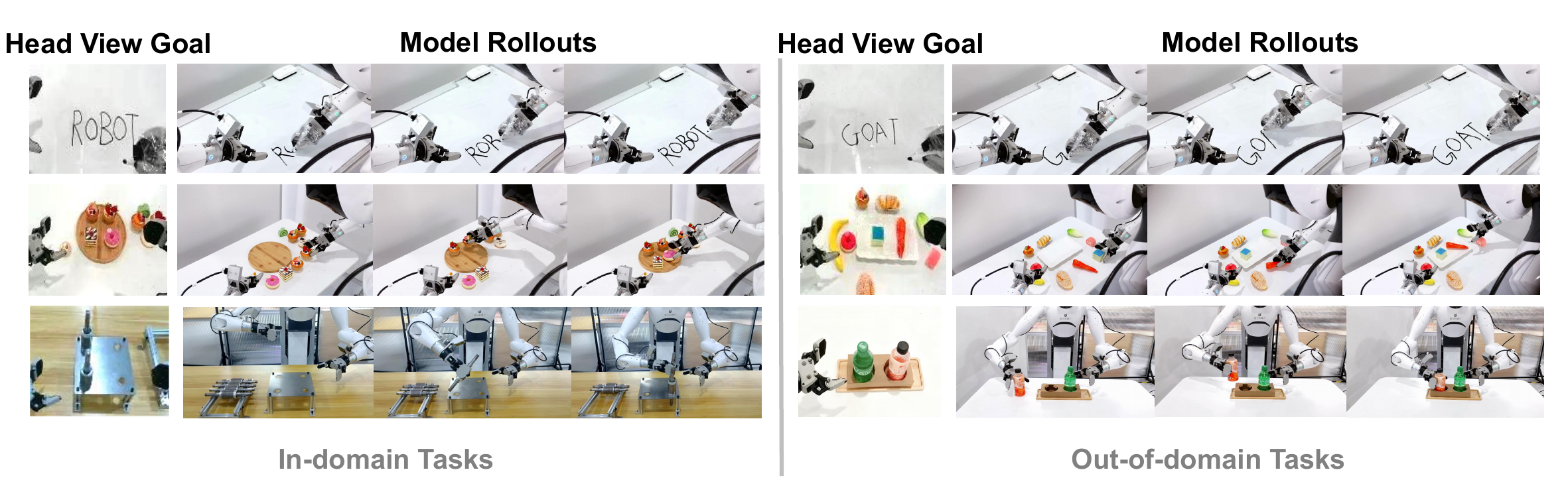}
    \vspace{-0.5cm}
    \caption{\textbf{Real World Evaluation.}
    This figure illustrates in-domain and out-of-domain test configurations for three real-world tasks: Whiteboard Word Writing, Dessert Plating, and Plug-In Operation. For each task, Head View Goal displays the target, while Model Rollouts shows the robot’s execution process; these setups are used to evaluate the model’s generalization ability using the success rate as the metric.}
    \label{fig:realtask}
\end{figure*}

\begin{table}[t]
  \centering
  \caption{\textbf{Comparison of Performance on Robotwin 2.0 Simulation Benchmark.}
  Act2Goal outperforms baselines in all Easy-mode tasks and 3 Hard-mode tasks, showing superior generalization.}
  \label{tab:robotwin2_performance_ext}
  \setlength{\tabcolsep}{1pt} 
  \begin{tabular*}{0.48\textwidth}{@{\extracolsep{\fill}} l l c c c c @{}}
    \toprule
    & Model/Task 
    & \makecell{Move \\ Can} 
    & \makecell{Pick \\ Bottles} 
    & \makecell{Place \\ Cup} 
    & \makecell{Place \\ Shoe} \\
    \midrule
    \multirow{4}{*}{\centering Easy}
    & DP-GC & 0.18 & 0.04 & 0.03 & 0.04 \\
    & $\pi_{0.5}$\textrm{-GC} & 0.54 & 0.13  & 0.16 & 0.30 \\
    & HyperGoalNet & 0.11 & 0.08 & 0.08 & 0.01 \\
    & Act2Goal & \textbf{0.62} & \textbf{0.80} & \textbf{0.64} & \textbf{0.52} \\
    \midrule
    \multirow{4}{*}{\centering Hard}
    & DP-GC & 0.00 & 0.00 & 0.00 & 0.00 \\
    & $\pi_{0.5}$\textrm{-GC} &\textbf{0.42} & 0.06  & 0.04 & 0.06 \\
    & HyperGoalNet & 0.00 & 0.00 & 0.00 & 0.00 \\
    & Act2Goal & 0.13 & \textbf{0.43} & \textbf{0.13} & \textbf{0.15} \\
    \bottomrule
  \end{tabular*}
  \vspace{-0.7cm}
\end{table}

\subsection{Generalization Capability Evaluation}
To comprehensively evaluate the generalization capability and versatility of Act2Goal after offline imitation training, we conducted extensive experiments in both simulated benchmarks and real-world robotic tasks. 
We selected four distinct tasks (Move can pot, Pick dual bottles, Place empty cup, Place shoe) from the Robotwin 2.0 simulation benchmark \cite{chen2025robotwin}. For each test episode in the simulator, the final goal image was obtained via  fixing the environment seed to ensure the reproducibility and consistency of the goal configuration across all test episodes, extracting it from successfully executed trajectories under this seed, and used as the goal condition for our model. The Robotwin 2.0 benchmark is particularly suitable for assessing generalization as it includes both seen scenarios (Easy mode) and highly challenging unseen scenarios (Hard mode).

We compare Act2Goal with several state-of-the-art policy architectures, including: (1) DP-GC, which conveys the SigLIP features of both the current observation and the goal image to a DiT-style Action Expert via cross-attention; (2) HyperGoalNet, a recent high-performing open-source goal-conditioned policy; and (3) the $\pi_{0.5}$-GC baseline, which uses a fixed language condition for all data while incorporating both a goal image and the current raw observation as visual inputs to $\pi_{0.5}$ \cite{ding2019goal,zhoutextit,intelligence2025pi05visionlanguageactionmodelopenworld, zhai2023sigmoid}. 

The results in Table ~\ref{tab:robotwin2_performance_ext} demonstrate that Act2Goal significantly outperforms all these baselines. Notably, it exhibits an overwhelming advantage in the Hard mode, particularly in unseen scenarios, highlighting its superior generalization capability.

\begin{table}[t]
  \centering
  \caption{\textbf{Comparison of Performance on Real-World Manipulation Tasks.}
  Act2Goal demonstrates remarkable performance across all real-world tasks, significantly outperforming all baseline methods. The robust performance in OOD settings strongly validates its superior generalization ability. All results are obtained without any online autonomous improvement, using only offline imitation learning.}
  \label{tab:real_task_performance_ext}
  \setlength{\tabcolsep}{1pt} 
  \begin{tabular*}{0.48\textwidth}{@{\extracolsep{\fill}} l l c c c @{}}
    \toprule
    & Model/Task
    & \makecell{Whiteboard \\ Word Writing}
    & \makecell{Dessert \\ Plating} 
    & \makecell{Plug-In \\ Operation} \\
    \midrule
    \multirow{4}{*}{\centering ID}
    & DP-GC & 0.00 & 0.10& 0.00 \\
    & $\pi_{0.5}$\textrm{-GC} & 0.23 & 0.18 & 0.00 \\
    & HyperGoalNet & 0.00 & 0.08 & 0.00 \\
    & Act2Goal & \textbf{0.93} & \textbf{0.75} & \textbf{0.45}  \\
    \midrule
    \multirow{4}{*}{\centering OOD}
    & DP-GC & 0.00 & 0.00 & 0.00 \\
    & $\pi_{0.5}$\textrm{-GC} &  0.20&  0.05& 0.00  \\
    & HyperGoalNet & 0.00 & 0.00 & 0.00 \\
    & Act2Goal & \textbf{0.90} & \textbf{0.48} & \textbf{0.30} \\
    \bottomrule
  \end{tabular*}
  \vspace{-0.7cm}
\end{table}

To further validate the generalization capability of Act2Goal in realistic scenarios, we conducted real-world experiments on an AgiBot Genie-01 robot using three challenging manipulation tasks. These tasks are characterized by objectives that are difficult to precisely specify via language, making them particularly suitable for goal-conditioned policies, and they require fine-grained control for successful execution. The data for these test tasks are part of the full dataset used for offline imitation training, and we did not perform any task-specific supervised fine-tuning (SFT). 
For each task, we designed both in-domain (ID) and out-of-domain (OOD) test configurations to thoroughly assess the model's generalization ability, using success rate as the evaluation metric. The three real-world tasks are designed as follows:
\begin{itemize}
    \item \textbf{Task 1: Whiteboard Word Writing.} The robot writes English words on a whiteboard. The ID test involves writing words seen in the training data (which includes 200 words), whereas the OOD test requires writing unseen words with novel character combinations, testing the model's compositional generalization.

    \item \textbf{Task 2: Dessert Plating.} The robot arranges desserts on a plate according to a visual example. The ID test uses dessert types, plate styles, and backgrounds seen during offline imitation training, while the OOD test introduces significant visual variations. This task evaluates the policy's goal-following capability under strong visual distractions.
    
    
    \item \textbf{Task 3: Plug-In Operation.} The robot performs insertion tasks. The ID test involves inserting a metal workpiece into a corresponding hole, as seen in training. The OOD test requires inserting a cylindrical drink bottle into a cup holder—an unseen task that assesses the model's skill transfer capability.
\end{itemize}


As summarized in Fig.~\ref{fig:realtask} and Table~\ref{tab:real_task_performance_ext}, Act2Goal demonstrates remarkable performance across all real-world tasks, significantly outperforming all baseline methods. Its robust performance in OOD settings strongly validates the superior generalization ability of the proposed approach.

\begin{figure}[t]

    \centering
    \includegraphics[width=\linewidth, trim=30 250 60 0, clip]{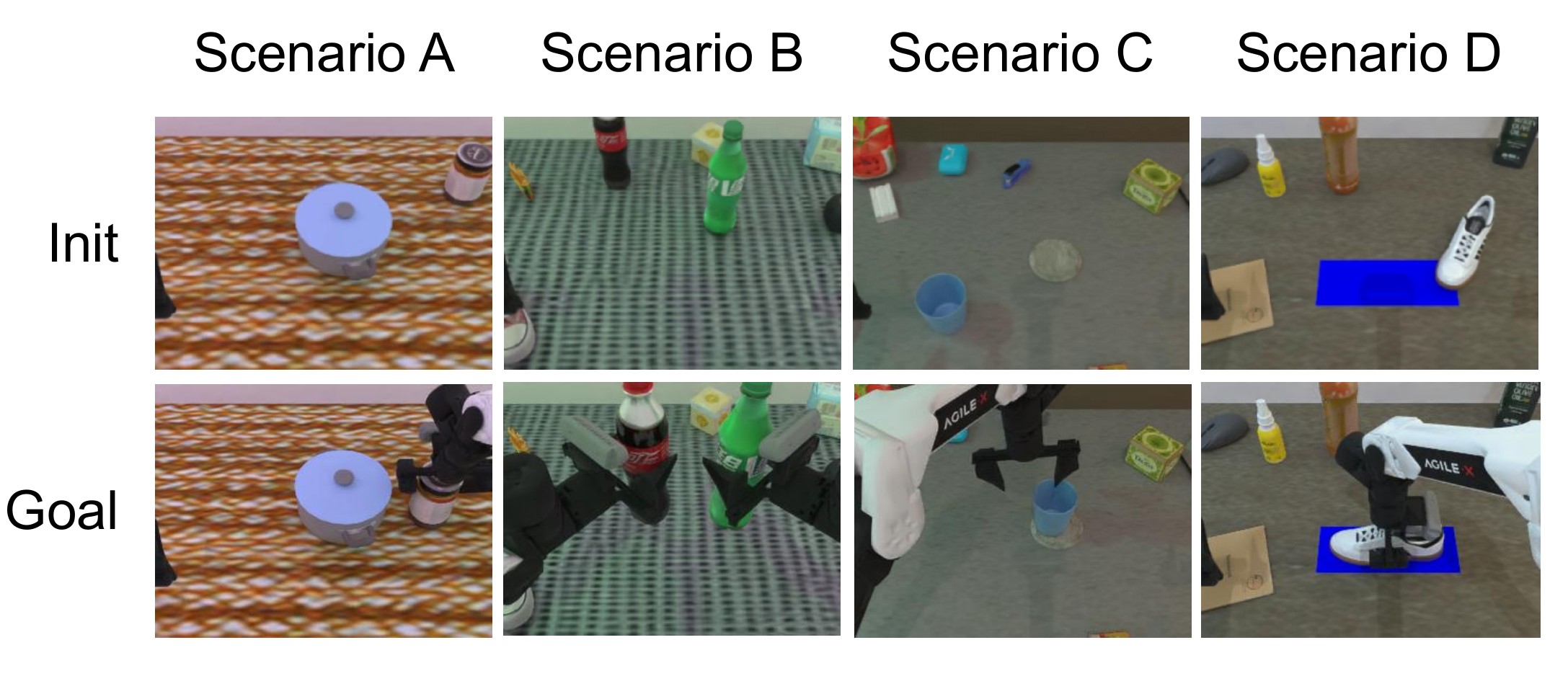}
    \vspace{-0.7cm}
    \caption{\textbf{Online Autonomous Improvement Scenarios.}
    This figure illustrates four OOD scenarios from the RoboTwin 2.0 benchmark, corresponding to the hard testing modes of Move Can Pot, Pick Dual Bottles, Place Empty Cup, and Place Shoe. These scenarios serve as the testbed for verifying the effectiveness of autonomous improvement.}
    \label{fig:online_train_scenario}
    \vspace{-0.3cm}
\end{figure}

\subsection{Analysis of Online Autonomous Improvement}
We validated the effectiveness of online autonomous improvement in both simulated and real-world environments, and systematically analyzed best practices for enhancing robot manipulation performance using Hindsight Experience Replay (HER).

\begin{figure}[t]
    \centering
    \includegraphics[width=\linewidth, trim=0 0 0 0, clip]{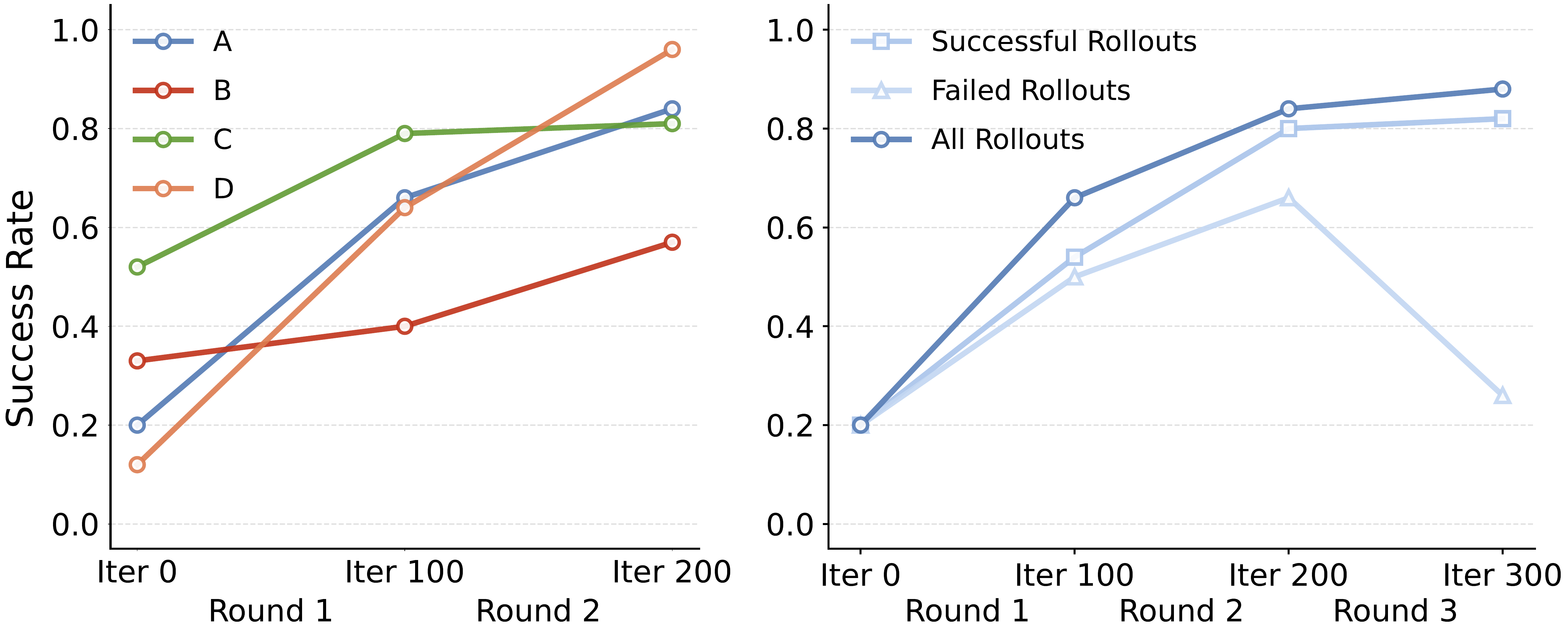}
    \caption{\textbf{Online Training Performance in Robotwin 2.0.} 
    This figure presents two key findings from Robotwin 2.0 simulations: 
    (Left) Multi-round success rates of four hard-mode scenarios, showing consistent improvement over 3 rounds before convergence. 
    (Right) Performance of three data selection strategies for rollouts: using all rollouts yields optimal results, while even failed-only rollouts enable noticeable improvement.}
    \label{fig:robotwin_online_train}
\end{figure}

In the Robotwin 2.0 simulation environment, we selected one hard-mode scenario from each of the four tasks described in Section 4.2 (shown in Fig.~\ref{fig:online_train_scenario}) for multi-round online training, evaluating the success rate after each round. As shown in the left part of Fig.~\ref{fig:robotwin_online_train}, the model’s performance improved consistently over approximately three rounds of training before converging, with a maximum success rate improvement of up to 8× compared to the pre-trained baseline. We also compared three data selection strategies for rollout usage: (1) using only successful rollouts, (2) using all rollouts regardless of success, and (3) using only failed rollouts. We found that utilizing all rollouts yielded the best performance. Notably, even using only failed rollouts still led to clear improvement, demonstrating that the HER-based strategy—extracting useful experience from failures—is effective for goal-conditioned robot manipulation. The corresponding results can be find in the right part of Fig.~\ref{fig:robotwin_online_train}.

\begin{figure}[t]

    \centering
    \includegraphics[width=\linewidth, trim=50 10 50 50, clip]{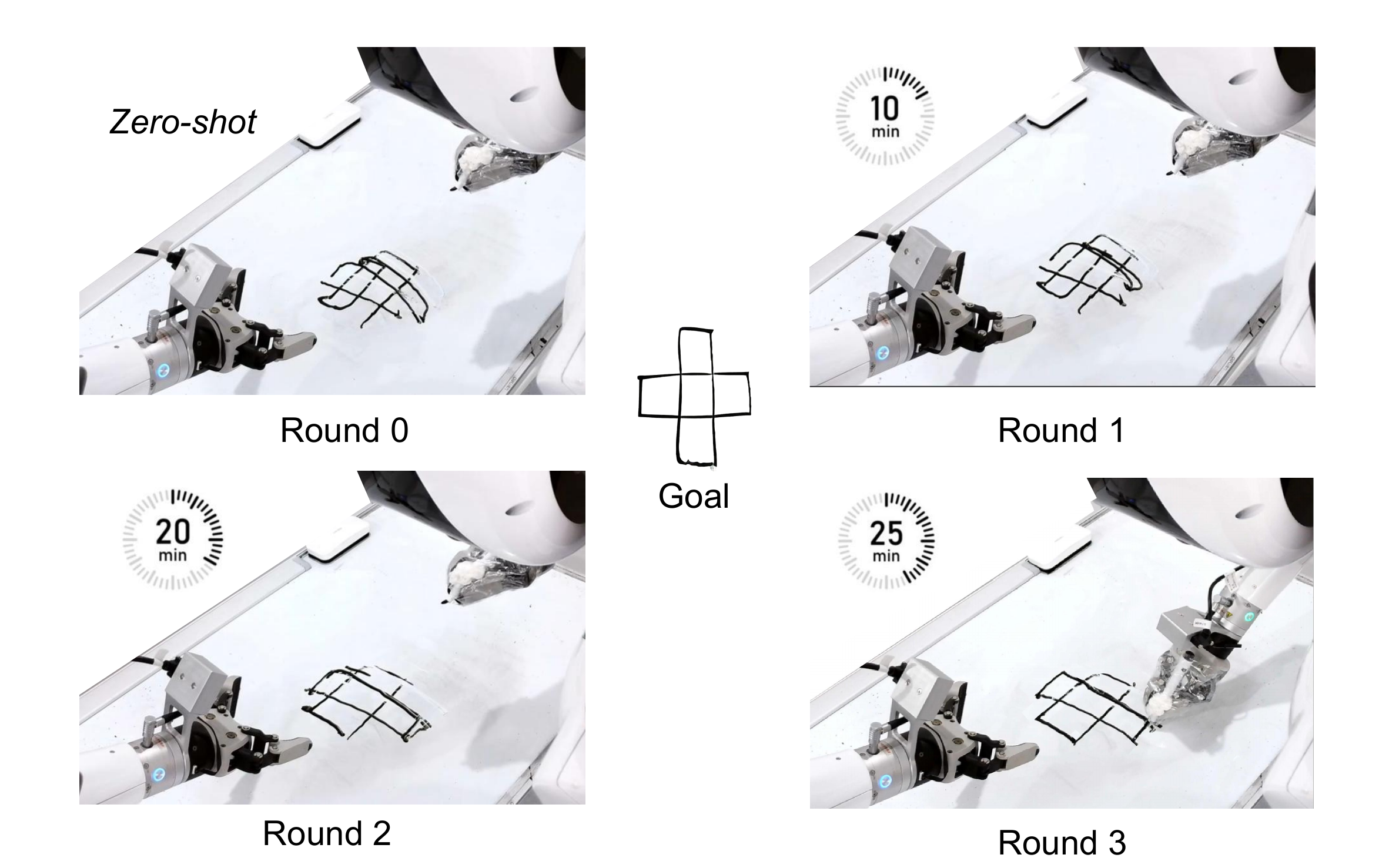}
    \caption{\textbf{Online Training Performance in Real-World Unseen Scenarios.}
    When tasked with drawing an unseen pattern, the model initially performs poorly. However, as online training progresses (from left to right, top to bottom), the drawing quality improves steadily.}
    \label{fig:draw_online_train}
    \vspace{-0.3cm}
\end{figure}

\begin{figure*}[t]
    \centering
    \includegraphics[width=\linewidth, trim=0 110 0 0, clip]{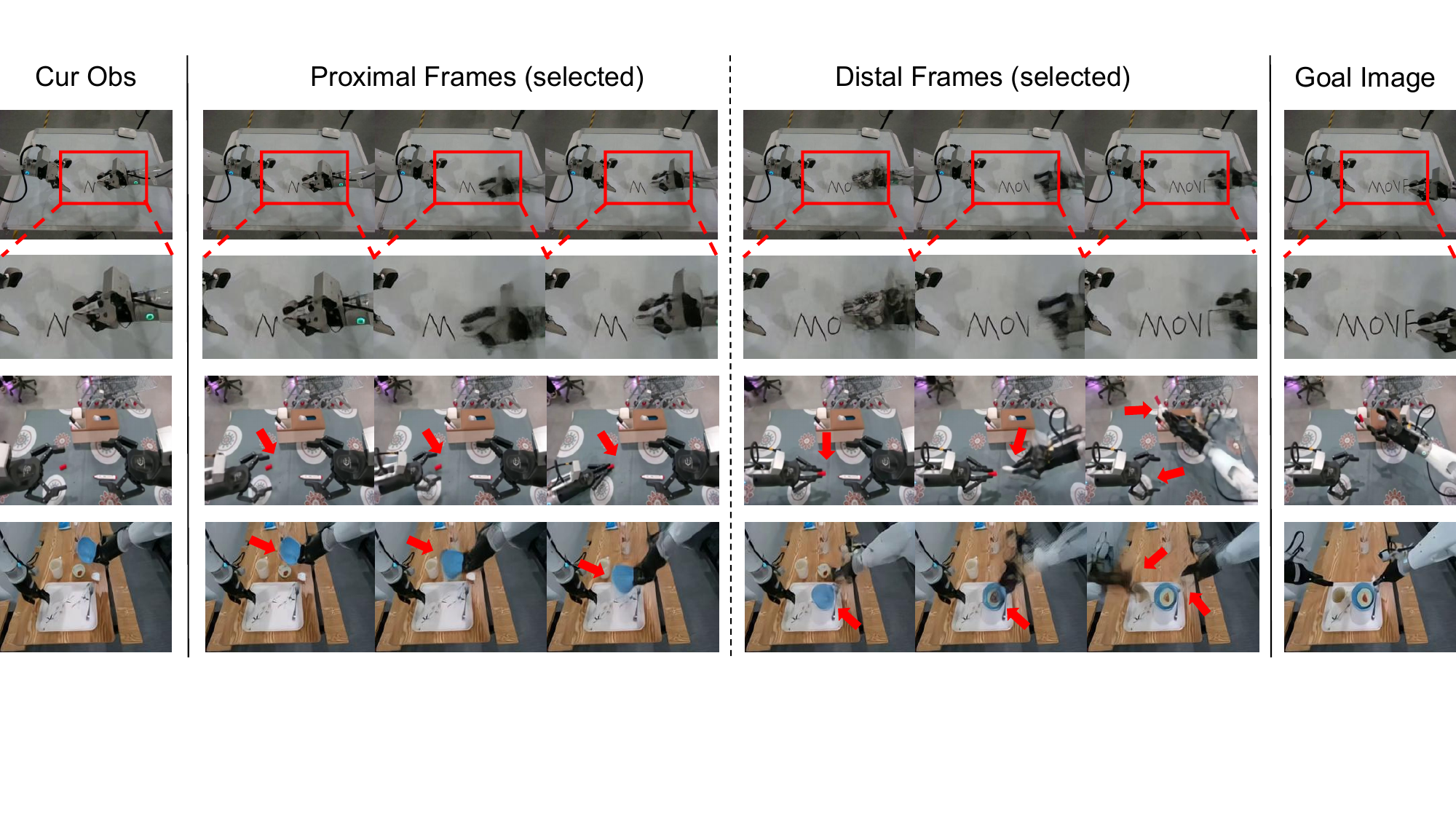}
    \caption{\textbf{Examples of generated videos.}
    This figure illustrates the generation capability of our goal-conditioned world model through three head-view video clips. The left and right sides show the current observation and the target goal, respectively. For each sequence, we uniformly sample three proximal and three distal frames. The second row provides a zoomed-in view of the red-highlighted regions in the first row. Red arrows in the third and fourth rows highlight the motion of key objects. The distal frames in the bottom row correspond to longer temporal horizons compared to those in the third row.}
    \label{fig:msth_vid}
\end{figure*}

In real-world experiments, we tasked the robot with whiteboard writing (Task 1 in Sec. 4.3). While the model exhibits strong generalization in whiteboard word writing, we assessed its capability to draw novel patterns that are substantially different from writing. As shown in Fig.~\ref{fig:draw_online_train}, the model's initial performance on this unfamiliar case was limited. Nevertheless, through online fine-tuning, it demonstrated consistent improvements in drawing quality in 15 minutes. On the OOD Plug-In task, the model also demonstrates a consistent increase in success rate throughout online training (from 0.30 to 0.90). Detailed information can be found in our project page.

\begin{table}[t]
  \centering
  \caption{\textbf{Effectiveness of MSTH.}
  Values denote the success rate of the real-world whiteboard word writing task. The word length definitions: Short (3 letters or fewer), Medium (4 to 6 letters), Long (7 letters or more).}
  \label{tab:msth_id_ood}
  \setlength{\tabcolsep}{1pt}  
  \begin{tabular*}{0.45\textwidth}{@{\extracolsep{\fill}} l l c c c @{}}
    \toprule
     & \textbf{Model} & \textbf{Short} & \textbf{Medium} & \textbf{Long} \\
    \midrule
    \multirow{2}{*}{\centering ID} 
    & w/o MSTH & 0.95 & 0.35 & 0.10 \\
    & w/ MSTH  & \textbf{0.95} & \textbf{0.90} & \textbf{0.90} \\ 
    \midrule
    \multirow{2}{*}{\centering OOD} 
    & w/o MSTH & 0.60 & 0.20 & 0.00 \\ 
    & w/ MSTH  & \textbf{0.93} & \textbf{0.90} & \textbf{0.88} \\
    \bottomrule
  \end{tabular*}
  \vspace{-0.5cm}
\end{table}

\subsection{Effectiveness of MSTH}

Fig.~\ref{fig:msth_vid} presents three representative examples of video clips generated by our goal-conditioned world model. As illustrated, the model is capable of producing both fine-grained proximal frames, which offer detailed short-term guidance essential for the action expert to generate accurate and executable actions, and sparser distal frames, which capture the long-horizon structure of the task and effectively guide the model toward the intended long-term goal. This hierarchical visual forecasting enables the model to reason across multiple temporal scales, facilitating both precise control and goal-directed planning.


We also conducted an ablation study to investigate the effectiveness of the proposed Multi-Scale Temporal Hashing (MSTH) mechanism in a real-world whiteboard word writing task, which requires long-horizon, visually grounded planning. As shown in Table~\ref{tab:msth_id_ood}, we compare the success rates across words of varying lengths—short (no more than 3 letters), medium (4–6 letters), and long (no fewer than 7 letters)—under both in-domain (ID) and out-of-domain (OOD) settings.

The MSTH-based policy demonstrates a substantial improvement over the baseline strategy that relies on fixed-horizon action chunking. While both methods perform comparably on short-horizon tasks (e.g., short words), the fixed-horizon baseline exhibits a drastic performance drop as the planning horizon increases, particularly in long and OOD scenarios. This degradation is primarily due to compounding goal misalignment over extended sequences, where the model struggles to maintain consistent intent.

In contrast, MSTH maintains high success rates across all word lengths and generalizes well to OOD words, indicating its robustness and effectiveness in handling long-horizon dependencies. By dynamically adjusting the temporal abstraction level and enabling better temporal anchoring to final goal, MSTH provides a simple yet principled mechanism for improving long-horizon goal-conditioned manipulation. These results highlight MSTH as a critical design component for scaling visual policy models to complex, temporally extended tasks.

\section{Conclusion} 
\label{sec:conclusion}

We propose Act2Goal, a goal-conditioned policy that combines a visual world model with multi-scale temporal control to tackle long-horizon manipulation. The model generates both fine-grained short-term executable control signals and coarse long-term trajectories, enabling effective planning and control.
Trained fully offline, Act2Goal achieves high success rates in seen tasks and shows strong generalization to unseen scenarios. To enhance adaptability, we incorporate Hindsight Experience Replay (HER) and LoRA-based finetuning, enabling efficient, reward-free online autonomous improvement within minutes.
Together, these components contribute to a scalable and generalizable visuomotor policy for open-ended, real-world manipulation tasks.

\bibliographystyle{unsrtnat}
\bibliography{references}

\begin{thebibliography}{47}
\providecommand{\natexlab}[1]{#1}
\providecommand{\url}[1]{\texttt{#1}}
\expandafter\ifx\csname urlstyle\endcsname\relax
  \providecommand{\doi}[1]{doi: #1}\else
  \providecommand{\doi}{doi: \begingroup \urlstyle{rm}\Url}\fi

\bibitem[Kaelbling(1993)]{kaelbling1993learning}
Leslie~Pack Kaelbling.
\newblock Learning to achieve goals.
\newblock In \emph{IJCAI}, volume~2, pages 1094--8, 1993.

\bibitem[Liu et~al.(2022)Liu, Zhu, and Zhang]{liu2022goal}
Minghuan Liu, Menghui Zhu, and Weinan Zhang.
\newblock Goal-conditioned reinforcement learning: Problems and solutions.
\newblock \emph{arXiv preprint arXiv:2201.08299}, 2022.

\bibitem[Ding et~al.(2019)Ding, Florensa, Abbeel, and Phielipp]{ding2019goal}
Yiming Ding, Carlos Florensa, Pieter Abbeel, and Mariano Phielipp.
\newblock Goal-conditioned imitation learning.
\newblock \emph{Advances in neural information processing systems}, 32, 2019.

\bibitem[Gong et~al.(2024)Gong, Feng, Xu, Ding, and Wang]{gong2024goal}
Xudong Gong, Dawei Feng, Kele Xu, Bo~Ding, and Huaimin Wang.
\newblock Goal-conditioned on-policy reinforcement learning.
\newblock \emph{Advances in neural information processing systems}, 37:\penalty0 45975--46001, 2024.

\bibitem[Black et~al.(2023)Black, Nakamoto, Atreya, Walke, Finn, Kumar, and Levine]{black2023zero}
Kevin Black, Mitsuhiko Nakamoto, Pranav Atreya, Homer Walke, Chelsea Finn, Aviral Kumar, and Sergey Levine.
\newblock Zero-shot robotic manipulation with pretrained image-editing diffusion models.
\newblock \emph{arXiv preprint arXiv:2310.10639}, 2023.

\bibitem[Tian et~al.(2024)Tian, Yang, Zeng, Wang, Lin, Dong, and Pang]{tian2024predictive}
Yang Tian, Sizhe Yang, Jia Zeng, Ping Wang, Dahua Lin, Hao Dong, and Jiangmiao Pang.
\newblock Predictive inverse dynamics models are scalable learners for robotic manipulation.
\newblock \emph{arXiv preprint arXiv:2412.15109}, 2024.

\bibitem[Reuss et~al.(2023)Reuss, Li, Jia, and Lioutikov]{reuss2023goal}
Moritz Reuss, Maximilian Li, Xiaogang Jia, and Rudolf Lioutikov.
\newblock Goal-conditioned imitation learning using score-based diffusion policies.
\newblock \emph{arXiv preprint arXiv:2304.02532}, 2023.

\bibitem[Liu et~al.(2024)Liu, Zhang, Li, Yan, Gao, Chen, Yuan, Huang, Sun, Gao, et~al.]{liu2024sora}
Yixin Liu, Kai Zhang, Yuan Li, Zhiling Yan, Chujie Gao, Ruoxi Chen, Zhengqing Yuan, Yue Huang, Hanchi Sun, Jianfeng Gao, et~al.
\newblock Sora: A review on background, technology, limitations, and opportunities of large vision models.
\newblock \emph{arXiv preprint arXiv:2402.17177}, 2024.

\bibitem[HaCohen et~al.(2024)HaCohen, Chiprut, Brazowski, Shalem, Moshe, Richardson, Levin, Shiran, Zabari, Gordon, et~al.]{hacohen2024ltx}
Yoav HaCohen, Nisan Chiprut, Benny Brazowski, Daniel Shalem, Dudu Moshe, Eitan Richardson, Eran Levin, Guy Shiran, Nir Zabari, Ori Gordon, et~al.
\newblock Ltx-video: Realtime video latent diffusion.
\newblock \emph{arXiv preprint arXiv:2501.00103}, 2024.

\bibitem[Agarwal et~al.(2025)Agarwal, Ali, Bala, Balaji, Barker, Cai, Chattopadhyay, Chen, Cui, Ding, et~al.]{agarwal2025cosmos}
Niket Agarwal, Arslan Ali, Maciej Bala, Yogesh Balaji, Erik Barker, Tiffany Cai, Prithvijit Chattopadhyay, Yongxin Chen, Yin Cui, Yifan Ding, et~al.
\newblock Cosmos world foundation model platform for physical ai.
\newblock \emph{arXiv preprint arXiv:2501.03575}, 2025.

\bibitem[Hu et~al.(2024)Hu, Guo, Wang, Chen, Wang, Zhang, Sreenath, Lu, and Chen]{hu2024video}
Yucheng Hu, Yanjiang Guo, Pengchao Wang, Xiaoyu Chen, Yen-Jen Wang, Jianke Zhang, Koushil Sreenath, Chaochao Lu, and Jianyu Chen.
\newblock Video prediction policy: A generalist robot policy with predictive visual representations.
\newblock \emph{arXiv preprint arXiv:2412.14803}, 2024.

\bibitem[Wen et~al.(2024)Wen, Lin, Zhu, Han, Xu, Zhao, and Liang]{wen2024vidman}
Youpeng Wen, Junfan Lin, Yi~Zhu, Jianhua Han, Hang Xu, Shen Zhao, and Xiaodan Liang.
\newblock Vidman: Exploiting implicit dynamics from video diffusion model for effective robot manipulation.
\newblock \emph{Advances in Neural Information Processing Systems}, 37:\penalty0 41051--41075, 2024.

\bibitem[Liao et~al.(2025)Liao, Zhou, Huang, Yang, Chen, Jiang, Hu, Cai, Liu, Luo, et~al.]{liao2025genie}
Yue Liao, Pengfei Zhou, Siyuan Huang, Donglin Yang, Shengcong Chen, Yuxin Jiang, Yue Hu, Jingbin Cai, Si~Liu, Jianlan Luo, et~al.
\newblock Genie envisioner: A unified world foundation platform for robotic manipulation.
\newblock \emph{arXiv preprint arXiv:2508.05635}, 2025.

\bibitem[Andrychowicz et~al.(2017)Andrychowicz, Wolski, Ray, Schneider, Fong, Welinder, McGrew, Tobin, Pieter~Abbeel, and Zaremba]{andrychowicz2017hindsight}
Marcin Andrychowicz, Filip Wolski, Alex Ray, Jonas Schneider, Rachel Fong, Peter Welinder, Bob McGrew, Josh Tobin, OpenAI Pieter~Abbeel, and Wojciech Zaremba.
\newblock Hindsight experience replay.
\newblock \emph{Advances in neural information processing systems}, 30, 2017.

\bibitem[Hu et~al.(2022)Hu, Shen, Wallis, Allen-Zhu, Li, Wang, Wang, Chen, et~al.]{hu2022lora}
Edward~J Hu, Yelong Shen, Phillip Wallis, Zeyuan Allen-Zhu, Yuanzhi Li, Shean Wang, Lu~Wang, Weizhu Chen, et~al.
\newblock Lora: Low-rank adaptation of large language models.
\newblock \emph{ICLR}, 1\penalty0 (2):\penalty0 3, 2022.

\bibitem[Jain and Unhelkar(2024)]{jain2024go}
Abhinav Jain and Vaibhav Unhelkar.
\newblock Go-dice: Goal-conditioned option-aware offline imitation learning via stationary distribution correction estimation.
\newblock In \emph{Proceedings of the AAAI conference on artificial intelligence}, volume~38, pages 12763--12772, 2024.

\bibitem[Kim et~al.(2024)Kim, Ohmura, and Kuniyoshi]{kim2024goal}
Heecheol Kim, Yoshiyuki Ohmura, and Yasuo Kuniyoshi.
\newblock Goal-conditioned dual-action imitation learning for dexterous dual-arm robot manipulation.
\newblock \emph{IEEE Transactions on Robotics}, 40:\penalty0 2287--2305, 2024.

\bibitem[Wang et~al.(2023)Wang, Fan, Sun, Zhang, Fei-Fei, Xu, Zhu, and Anandkumar]{wang2023mimicplay}
Chen Wang, Linxi Fan, Jiankai Sun, Ruohan Zhang, Li~Fei-Fei, Danfei Xu, Yuke Zhu, and Anima Anandkumar.
\newblock Mimicplay: Long-horizon imitation learning by watching human play.
\newblock \emph{arXiv preprint arXiv:2302.12422}, 2023.

\bibitem[Zhou et~al.()Zhou, Yao, Luo, Zhou, and Yang]{zhoutextit}
Pei Zhou, Wanting Yao, Qian Luo, Xunzhe Zhou, and Yanchao Yang.
\newblock Hyper-goalnet: Goal-conditioned manipulation policy learning with hypernetworks.
\newblock In \emph{The Thirty-ninth Annual Conference on Neural Information Processing Systems}.

\bibitem[Wen et~al.(2023)Wen, Lin, So, Chen, Dou, Gao, and Abbeel]{wen2023any}
Chuan Wen, Xingyu Lin, John So, Kai Chen, Qi~Dou, Yang Gao, and Pieter Abbeel.
\newblock Any-point trajectory modeling for policy learning.
\newblock \emph{arXiv preprint arXiv:2401.00025}, 2023.

\bibitem[Bharadhwaj et~al.(2024)Bharadhwaj, Mottaghi, Gupta, and Tulsiani]{bharadhwaj2024track2act}
Homanga Bharadhwaj, Roozbeh Mottaghi, Abhinav Gupta, and Shubham Tulsiani.
\newblock Track2act: Predicting point tracks from internet videos enables generalizable robot manipulation.
\newblock In \emph{European Conference on Computer Vision}, pages 306--324. Springer, 2024.

\bibitem[Yin et~al.(2025)Yin, Yang, and Abbeel]{yin2025object}
Zhao-Heng Yin, Sherry Yang, and Pieter Abbeel.
\newblock Object-centric 3d motion field for robot learning from human videos.
\newblock \emph{arXiv preprint arXiv:2506.04227}, 2025.

\bibitem[Davidson et~al.(2025)Davidson, Todd, Togelius, Gureckis, and Lake]{davidson2025goals}
Guy Davidson, Graham Todd, Julian Togelius, Todd~M Gureckis, and Brenden~M Lake.
\newblock Goals as reward-producing programs.
\newblock \emph{Nature Machine Intelligence}, 7\penalty0 (2):\penalty0 205--220, 2025.

\bibitem[Zhang et~al.(2025)Zhang, Hu, Qiao, Zhang, Qin, Li, Liu, Kong, Liu, and Ma]{zhang2025chain}
Wenbo Zhang, Tianrun Hu, Yanyuan Qiao, Hanbo Zhang, Yuchu Qin, Yang Li, Jiajun Liu, Tao Kong, Lingqiao Liu, and Xiao Ma.
\newblock Chain-of-action: Trajectory autoregressive modeling for robotic manipulation.
\newblock \emph{arXiv preprint arXiv:2506.09990}, 2025.

\bibitem[Ying et~al.(2025)Ying, Collins, Sharma, Colas, Zhao, Weller, Tavares, Isola, Gershman, Andreas, et~al.]{ying2025assessing}
Lance Ying, Katherine~M Collins, Prafull Sharma, Cedric Colas, Kaiya~Ivy Zhao, Adrian Weller, Zenna Tavares, Phillip Isola, Samuel~J Gershman, Jacob~D Andreas, et~al.
\newblock Assessing adaptive world models in machines with novel games.
\newblock \emph{arXiv preprint arXiv:2507.12821}, 2025.

\bibitem[Ha and Schmidhuber(2018)]{ha2018world}
David Ha and J{\"u}rgen Schmidhuber.
\newblock World models.
\newblock \emph{arXiv preprint arXiv:1803.10122}, 2\penalty0 (3), 2018.

\bibitem[Hafner et~al.(2019)Hafner, Lillicrap, Ba, and Norouzi]{hafner2019dream}
Danijar Hafner, Timothy Lillicrap, Jimmy Ba, and Mohammad Norouzi.
\newblock Dream to control: Learning behaviors by latent imagination.
\newblock \emph{arXiv preprint arXiv:1912.01603}, 2019.

\bibitem[Hafner et~al.(2023)Hafner, Pasukonis, Ba, and Lillicrap]{hafner2023mastering}
Danijar Hafner, Jurgis Pasukonis, Jimmy Ba, and Timothy Lillicrap.
\newblock Mastering diverse domains through world models.
\newblock \emph{arXiv preprint arXiv:2301.04104}, 2023.

\bibitem[Zhao et~al.(2025)Zhao, Zeng, Zhuang, Zhao, Xue, Wang, Zhao, Li, Li, Huang, et~al.]{zhao2025high}
Haoyu Zhao, Cheng Zeng, Linghao Zhuang, Yaxi Zhao, Shengke Xue, Hao Wang, Xingyue Zhao, Zhongyu Li, Kehan Li, Siteng Huang, et~al.
\newblock High-fidelity simulated data generation for real-world zero-shot robotic manipulation learning with gaussian splatting.
\newblock \emph{arXiv preprint arXiv:2510.10637}, 2025.

\bibitem[Liu et~al.(2025)Liu, Wang, Zhao, Li, Qin, Qiu, Zhu, Huang, and Su]{liu2025robotransfer}
Liu Liu, Xiaofeng Wang, Guosheng Zhao, Keyu Li, Wenkang Qin, Jiaxiong Qiu, Zheng Zhu, Guan Huang, and Zhizhong Su.
\newblock Robotransfer: Geometry-consistent video diffusion for robotic visual policy transfer.
\newblock \emph{arXiv preprint arXiv:2505.23171}, 2025.

\bibitem[Jiang et~al.(2025)Jiang, Chen, Huang, Chen, Zhou, Liao, He, Liu, Li, Yao, et~al.]{jiang2025enerverse}
Yuxin Jiang, Shengcong Chen, Siyuan Huang, Liliang Chen, Pengfei Zhou, Yue Liao, Xindong He, Chiming Liu, Hongsheng Li, Maoqing Yao, et~al.
\newblock Enerverse-ac: Envisioning embodied environments with action condition.
\newblock \emph{arXiv preprint arXiv:2505.09723}, 2025.

\bibitem[Guo et~al.(2025)Guo, Shi, Chen, and Finn]{guo2025ctrl}
Yanjiang Guo, Lucy~Xiaoyang Shi, Jianyu Chen, and Chelsea Finn.
\newblock Ctrl-world: A controllable generative world model for robot manipulation.
\newblock \emph{arXiv preprint arXiv:2510.10125}, 2025.

\bibitem[Huang et~al.(2025)Huang, Chen, Zhou, Chen, Jiang, Hu, Liao, Gao, Li, Yao, et~al.]{huang2025enerverse}
Siyuan Huang, Liliang Chen, Pengfei Zhou, Shengcong Chen, Zhengkai Jiang, Yue Hu, Yue Liao, Peng Gao, Hongsheng Li, Maoqing Yao, et~al.
\newblock Enerverse: Envisioning embodied future space for robotics manipulation.
\newblock \emph{arXiv preprint arXiv:2501.01895}, 2025.

\bibitem[Liang et~al.(2025)Liang, Tokmakov, Liu, Sudhakar, Shah, Ambrus, and Vondrick]{liang2025video}
Junbang Liang, Pavel Tokmakov, Ruoshi Liu, Sruthi Sudhakar, Paarth Shah, Rares Ambrus, and Carl Vondrick.
\newblock Video generators are robot policies.
\newblock \emph{arXiv preprint arXiv:2508.00795}, 2025.

\bibitem[Cen et~al.(2025)Cen, Yu, Yuan, Jiang, Huang, Guo, Li, Song, Luo, Wang, et~al.]{cen2025worldvla}
Jun Cen, Chaohui Yu, Hangjie Yuan, Yuming Jiang, Siteng Huang, Jiayan Guo, Xin Li, Yibing Song, Hao Luo, Fan Wang, et~al.
\newblock Worldvla: Towards autoregressive action world model.
\newblock \emph{arXiv preprint arXiv:2506.21539}, 2025.

\bibitem[Ross et~al.(2011)Ross, Gordon, and Bagnell]{ross2011reduction}
St{\'e}phane Ross, Geoffrey Gordon, and Drew Bagnell.
\newblock A reduction of imitation learning and structured prediction to no-regret online learning.
\newblock In \emph{Proceedings of the fourteenth international conference on artificial intelligence and statistics}, pages 627--635. JMLR Workshop and Conference Proceedings, 2011.

\bibitem[Kelly et~al.(2019)Kelly, Sidrane, Driggs-Campbell, and Kochenderfer]{kelly2019hg}
Michael Kelly, Chelsea Sidrane, Katherine Driggs-Campbell, and Mykel~J Kochenderfer.
\newblock Hg-dagger: Interactive imitation learning with human experts.
\newblock In \emph{2019 International Conference on Robotics and Automation (ICRA)}, pages 8077--8083. IEEE, 2019.

\bibitem[Luo et~al.(2025)Luo, Xu, Wu, and Levine]{luo2025precise}
Jianlan Luo, Charles Xu, Jeffrey Wu, and Sergey Levine.
\newblock Precise and dexterous robotic manipulation via human-in-the-loop reinforcement learning.
\newblock \emph{Science Robotics}, 10\penalty0 (105):\penalty0 eads5033, 2025.

\bibitem[Shah et~al.(2025)Shah, Liu, Wang, Jiang, Kumar, Seo, Mart{\'\i}n-Mart{\'\i}n, and Zhu]{shah2025mimicdroid}
Rutav Shah, Shuijing Liu, Qi~Wang, Zhenyu Jiang, Sateesh Kumar, Mingyo Seo, Roberto Mart{\'\i}n-Mart{\'\i}n, and Yuke Zhu.
\newblock Mimicdroid: In-context learning for humanoid robot manipulation from human play videos.
\newblock \emph{arXiv preprint arXiv:2509.09769}, 2025.

\bibitem[Sridhar et~al.(2025)Sridhar, Dutta, Jayaraman, and Lee]{sridhar2025ricl}
Kaustubh Sridhar, Souradeep Dutta, Dinesh Jayaraman, and Insup Lee.
\newblock Ricl: Adding in-context adaptability to pre-trained vision-language-action models.
\newblock \emph{arXiv preprint arXiv:2508.02062}, 2025.

\bibitem[Yang et~al.(2021)Yang, Fang, Han, Du, Luo, and Li]{yang2021mher}
Rui Yang, Meng Fang, Lei Han, Yali Du, Feng Luo, and Xiu Li.
\newblock Mher: Model-based hindsight experience replay.
\newblock \emph{arXiv preprint arXiv:2107.00306}, 2021.

\bibitem[Schramm et~al.(2023)Schramm, Deng, Granados, and Boularias]{schramm2023usher}
Liam Schramm, Yunfu Deng, Edgar Granados, and Abdeslam Boularias.
\newblock Usher: Unbiased sampling for hindsight experience replay.
\newblock In \emph{Conference on Robot Learning}, pages 2073--2082. PMLR, 2023.

\bibitem[Luo et~al.(2023)Luo, Wang, Dong, Zhang, Cheng, Sun, and Song]{luo2023relay}
Yongle Luo, Yuxin Wang, Kun Dong, Qiang Zhang, Erkang Cheng, Zhiyong Sun, and Bo~Song.
\newblock Relay hindsight experience replay: Self-guided continual reinforcement learning for sequential object manipulation tasks with sparse rewards.
\newblock \emph{Neurocomputing}, 557:\penalty0 126620, 2023.

\bibitem[Chen et~al.(2025)Chen, Chen, Chen, Cai, Liu, Li, Liang, Lin, Ge, Gu, et~al.]{chen2025robotwin}
Tianxing Chen, Zanxin Chen, Baijun Chen, Zijian Cai, Yibin Liu, Zixuan Li, Qiwei Liang, Xianliang Lin, Yiheng Ge, Zhenyu Gu, et~al.
\newblock Robotwin 2.0: A scalable data generator and benchmark with strong domain randomization for robust bimanual robotic manipulation.
\newblock \emph{arXiv preprint arXiv:2506.18088}, 2025.

\bibitem[Intelligence et~al.(2025)Intelligence, Black, Brown, Darpinian, Dhabalia, Driess, Esmail, Equi, Finn, Fusai, Galliker, Ghosh, Groom, Hausman, Ichter, Jakubczak, Jones, Ke, LeBlanc, Levine, Li-Bell, Mothukuri, Nair, Pertsch, Ren, Shi, Smith, Springenberg, Stachowicz, Tanner, Vuong, Walke, Walling, Wang, Yu, and Zhilinsky]{intelligence2025pi05visionlanguageactionmodelopenworld}
Physical Intelligence, Kevin Black, Noah Brown, James Darpinian, Karan Dhabalia, Danny Driess, Adnan Esmail, Michael Equi, Chelsea Finn, Niccolo Fusai, Manuel~Y. Galliker, Dibya Ghosh, Lachy Groom, Karol Hausman, Brian Ichter, Szymon Jakubczak, Tim Jones, Liyiming Ke, Devin LeBlanc, Sergey Levine, Adrian Li-Bell, Mohith Mothukuri, Suraj Nair, Karl Pertsch, Allen~Z. Ren, Lucy~Xiaoyang Shi, Laura Smith, Jost~Tobias Springenberg, Kyle Stachowicz, James Tanner, Quan Vuong, Homer Walke, Anna Walling, Haohuan Wang, Lili Yu, and Ury Zhilinsky.
\newblock $\pi_{0.5}$: a vision-language-action model with open-world generalization, 2025.
\newblock URL \url{https://arxiv.org/abs/2504.16054}.

\bibitem[Zhai et~al.(2023)Zhai, Mustafa, Kolesnikov, and Beyer]{zhai2023sigmoid}
Xiaohua Zhai, Basil Mustafa, Alexander Kolesnikov, and Lucas Beyer.
\newblock Sigmoid loss for language image pre-training.
\newblock In \emph{Proceedings of the IEEE/CVF international conference on computer vision}, pages 11975--11986, 2023.

\bibitem[Bu et~al.(2025)Bu, Cai, Chen, Cui, Ding, Feng, Gao, He, Hu, Huang, et~al.]{bu2025agibot}
Qingwen Bu, Jisong Cai, Li~Chen, Xiuqi Cui, Yan Ding, Siyuan Feng, Shenyuan Gao, Xindong He, Xuan Hu, Xu~Huang, et~al.
\newblock Agibot world colosseo: A large-scale manipulation platform for scalable and intelligent embodied systems.
\newblock \emph{arXiv preprint arXiv:2503.06669}, 2025.

\end{thebibliography}

\clearpage
\section*{Appendix}

\subsection{Implementation Details}

\textbf{Model.} The goal-conditioned world model predicts 4 latent frames (2 proximal, 2 distal), decoded by a 3D VAE into 9 proximal and 9 distal visual frames. The action expert outputs 54 proximal actions (execute 50) and 9 distal actions (for guidance only; not executed). At inference, only actions are generated.

\textbf{Training.} We train the model on the AgiBot World dataset and a small proprietary dataset \cite{bu2025agibot}. Stage 1 fine-tunes a pre-trained 1.6B-parameter Genie Envisioner (7×24 hours, 16×A800). Stage 2 performs end-to-end behavioral cloning (48 hours, 16×A800).

\textbf{Deployment and Online Learning.} The policy is deployed on an AgiBot Genie-01 robot with an NVIDIA RTX 4090. Online autonomous improvement fine-tunes only the LoRA layers (rank 64) using a replay buffer of size 20. Each training round (10 epochs) takes ~5 minutes including rollout, backpropgation and environment resetting. Inference latency is 200ms for 50 executable actions.

\subsection{Real World Task Setup}

\textbf{Writing and Painting Tasks.} For the tasks of writing and drawing on a whiteboard, the robot gripper was first manually positioned to grasp the marker before task execution. During extended writing trials, the smooth surface of the marker occasionally caused it to slip from the gripper. To ensure reliability and reduce resetting time during batch testing, we used tape to further secure the marker in place.

\textbf{Bearing Insertion Task.} In the demonstrated bearing insertion task, each bearing weighs over 2 kg, with a base diameter of approximately 1 cm. The bearing is inserted into a hole with a diameter of about 1.5 cm.

\textbf{Dessert Plating Task}. To ensure experimental repeatability, we used silicone-made toy desserts for the dessert plating task.

\subsection{Testing Metric}
We employ task execution success rate as the performance metric for evaluating the models. Specifically, the real‑world success rates reported in Section IV are obtained by manually labeling success or failure over 40 model rollouts per experiment. The success rates in simulation are automatically computed from 90 rollouts. For models in Section IV.B that underwent autonomous improvement, we saved the model weights after each training round and evaluated them individually once the autonomous improvement process was completed.

\section*{Acknowledgments}
We gratefully acknowledge the technical support and insightful discussions provided by Yifei Wei, Xindong He, Jingbin Cai, Yuxin Jiang, Siyuan Huang, Qi Qin, Yue Liao, Qi Lv, Sukai Wang, Ruofei Niu, Xiongfeng Cai, Hanwen Shi and Xuan Hu. We also thank Haoyu Cao, Cheng Jing, Rui Wu, Dan Liu and Jia Zeng for their diligent work in data preparation. 

\end{document}